\documentclass{article}
\usepackage{graphicx}
\usepackage{amsmath,amssymb}
\usepackage{multirow}
\usepackage{booktabs}
\usepackage{graphicx}
\usepackage{enumitem}
\usepackage{placeins}
\usepackage{array}
\usepackage{enumitem}

\usepackage[main, final]{neurips_2026_arxiv}

\usepackage{authblk}
\usepackage[utf8]{inputenc} 
\usepackage[T1]{fontenc}    
\usepackage{hyperref}       
\usepackage{url}            
\usepackage{booktabs}       
\usepackage{amsfonts}       
\usepackage{nicefrac}       
\usepackage{microtype}      
\usepackage{xcolor}         

\newcommand{\vc}[3]{\shortstack[c]{#1 \\ {\scriptsize [#2, #3]}}}
\title{Beyond ViT Tokens: Masked-Diffusion Pretrained Convolutional Pathology Foundation Model for Cell-Level Dense Prediction}

\author[${\ast}$,1]{\textbf{Weiming Chen}\thanks{Contributed equally}}

\author[$\boldsymbol{\ast}$,1,3]{\textbf{Xitong Ling}}
\author[2]{\textbf{Zhenyang Cai}}
\author[2]{\textbf{Xidong Wang}}
\author[1]{\textbf{Jiawen Li}}
\author[1]{\textbf{Tian Guan}}
\author[$\dagger$,2]{\textbf{Benyou Wang}}
\author[$\dagger$,1,3]{\textbf{Yonghong He}\thanks{Corresponding authors}}

\affil[1]{%
  \textup{Tsinghua Shenzhen International Graduate School, Tsinghua University}}
\affil[2]{%
  \textup{The Chinese University of Hong Kong, ShenZhen }}
\affil[3]{%
  \textup{Research Institute of Tsinghua, Pearl River Delta}}

\date{} 

\definecolor{darkgreen}{rgb}{0.0,0.5,0.0}
\begin{document}

\maketitle

\begin{abstract}
  Cell-level dense prediction is central to computational pathology, but remains challenging due to fine-grained histological structures, strong domain shifts, and costly dense annotations. Existing ViT-based pathology foundation models rely on patch tokenization, which can disrupt spatial continuity and weaken local morphological details needed for cell-level prediction. To address this, we propose Masked-Diffusion Convolutional Foundation Models, termed \textbf{C}onvNeXt \textbf{M}asked-\textbf{D}iffusion (\textbf{CMD}), a self-supervised convolutional generative pretraining framework for dense pathology representation learning. CMD uses a fully convolutional ConvNeXt-UNet backbone, performs masked-diffusion pretraining in pixel space, and incorporates frozen pathology foundation model features through adaptive normalization.  Experimental results demonstrate that CMD consistently outperforms existing ViT-based pathology foundation models and even surpasses state-of-the-art end-to-end segmentation methods while fine-tuning only a small number of task-specific parameters across multiple pathology dense prediction tasks. The advantage is particularly pronounced under limited annotation settings, where CMD exhibits stronger robustness and generalization ability. Our findings suggest that purely convolutional architectures can also serve as competitive pathology foundation models for cell-level dense prediction, achieving leading performance within the current ViT-dominated paradigm and providing a scalable, high-performance solution that better preserves histological structural priors for fine-grained pathology understanding. 
\end{abstract}

\section{Introduction}

Cell-level dense prediction is central to computational pathology, enabling nuclear segmentation, inflammatory cell detection, and tissue microenvironment analysis. This task remains difficult because pathological structures are often tiny, morphologically diverse, and separated by ambiguous boundaries, while dense annotations are costly and domain shifts across stains, scanners, tissues, and institutions are substantial.

Recent pathology foundation models largely build on ViT architectures. Although effective for high-level recognition, ViT-style patch tokenization is not naturally suited to cell-level dense prediction: fixed-size patches may disrupt continuous histological structures and lose intra-patch morphology, texture, and boundary details that are critical for pixel- or instance-level prediction. This suggests that convolutional networks, with their locality and spatial continuity biases, are a more suitable architectural choice for fine-grained pathology representation learning.

However, convolutional networks still lack a scalable pretraining paradigm comparable to masked image modeling for ViTs. Inspired by recent generative models, which demonstrate a strong ability to learn high-fidelity image representations through reconstruction and synthesis, we ask whether generative modeling can serve as an effective self-supervised pretraining recipe for convolutional pathology foundation models.

To this end, we propose Masked-Diffusion Convolutional Foundation Models, termed \textbf{C}onvNeXt \textbf{M}asked-\textbf{D}iffusion (\textbf{CMD}), a generative self-supervised pretraining framework for cell-level dense prediction in pathology. CMD performs masked-diffusion pretraining in pixel space with a convolutional architecture, preserving local spatial continuity while learning morphology-aware representations. We instantiate CMD with a ConvNeXt-UNet backbone and condition the diffusion process on pathology foundation model features, combining semantic priors with fine-grained spatial reconstruction.

Across multiple cell-level pathology benchmarks, including multi-dataset training, few-shot adaptation, and scaling settings, CMD consistently outperforms ViT-based pathology foundation models and state-of-the-art end-to-end segmentation networks.

Our contributions are summarized as follows:
\begin{enumerate}[
    label=\Roman*.,
    leftmargin=1.5em,
    itemindent=0em,
    labelsep=0.5em,
    itemsep=2pt,
    topsep=2pt
]
\item We propose \textbf{CMD}, a self-supervised Masked-Diffusion Convolutional Foundation Model for learning cell-level dense representations in pathology.
\item We systematically design the key components of the framework, including the masked-diffusion objective, ConvNeXt-UNet backbone, pixel-space modeling, and pathology foundation model conditioning.
\item We validate the generalizability of the learned representations across multiple datasets, few-shot scenarios, and scaling regimes, with visualizations showing localized, cell-aware dense features.
\end{enumerate}

\section{Related Work}

\paragraph{Pathology Foundation Models.}
Pathology foundation models (PFMs) provide transferable representations for computational pathology. Most existing PFMs follow vision-only self-supervised pretraining with ViT backbones and objectives such as DINOv2~\cite{oquab2023dinov2}, iBOT~\cite{zhou2021ibot}, or masked image modeling~\cite{he2022masked}, and are later paired with MIL-style aggregators~\cite{lu2021data,ling2024agent,luo2025nnmil} for WSI-level tasks. Representative models include UNI/UNI2~\cite{chen2024towards}, Virchow/Virchow2~\cite{vorontsov2024foundation,zimmermann2024virchow2}, PathOrchestra~\cite{yan2025pathorchestra}, Phikon~\cite{Filiot2023ScalingSSLforHistoWithMIM,filiot2024phikon}, Prov-GigaPath~\cite{xu2024whole}, Hibou~\cite{nechaev2024hibou}, Kaiko~\cite{aben2024towards}, Digepath~\cite{zhu2025subspecialty}, StainNet~\cite{li2025stainnet}, GPFM~\cite{ma2025generalizable}, Midnight-12k~\cite{KDK2025} and GenBio-PathFM~\cite{kapse2026genbio}. Another line uses vision--language pretraining to align pathology images with reports or biomedical text, such as PLIP~\cite{huang2023visual}, CONCH~\cite{lu2024visual,ding2025multimodal} and MUSK~\cite{xiang2025vision}. While effective for global image- or slide-level semantics, ViT-based PFMs may lose fine local continuity due to patch tokenization, limiting their suitability for cell-level dense prediction.
\paragraph{Dense Prediction in Pathology.}
Dense prediction~\cite{ronneberger2015u} supports pixel- or region-level analysis of nuclei, glands, tumor regions, immune cells, and tissue microenvironment components~\cite{liu2024panoptic}. Existing methods often adapt natural-image segmentation architectures to pathology. TransUNet~\cite{chen2021transunet} combines CNN local features with Transformer global context, ViT-Adapter~\cite{chen2022vision} adapts pretrained ViTs through multi-scale modules. These methods improve supervised dense prediction, but typically require end-to-end training with dense annotations.
\paragraph{Generative and Convolutional Pretraining.}
Generative pretraining learns representations by reconstructing or synthesizing image content. Masked image modeling is widely used for ViT pretraining~\cite{he2022masked}, while diffusion models learn visual structure through denoising objectives. Masked diffusion~\cite{pan2023masked} further views diffusion as time-conditioned reconstruction, suggesting that the corruption process can be designed for representation learning rather than image synthesis alone. In parallel, modern convolutional architectures such as ConvNeXt~\cite{liu2022convnet} and ConvNeXt V2~\cite{woo2023convnext} provide strong locality bias and efficient multi-scale feature extraction, making convolutional pretraining an important alternative to token-based visual representation learning.

\section{Method}
\label{headings}

We develop CMD by addressing five key design choices: the pretraining objective, backbone architecture, representation space, foundation-model conditioning, and downstream feature extraction. This section first summarizes the overall pipeline.

\subsection{Overview: Frozen Generative Pretraining for Cell-Level Dense Prediction}

CMD aims to learn a reusable generative pathology representation rather than another task-specific supervised segmentor. Given unlabeled pathology image patches, we pretrain a ConvNeXt masked-diffusion model in a self-supervised manner, following the masked diffusion formulation~\cite{pan2023masked}. The model takes partially masked patches as input and learns to recover the clean image content through a diffusion denoising objective, encouraging it to capture local morphology, texture continuity, and cellular boundaries without dense annotations.

After pretraining, we discard the sampling process and use the pretrained diffusion network as a frozen dense feature extractor. Multi-scale feature maps are extracted from intermediate decoder blocks, providing both fine spatial details and contextual information for downstream cell-level dense prediction. For each task, only a lightweight prediction head is trained, while the pretrained generative representation remains fixed.

\begin{figure}[t]
    \centering
    \includegraphics[width=\linewidth]{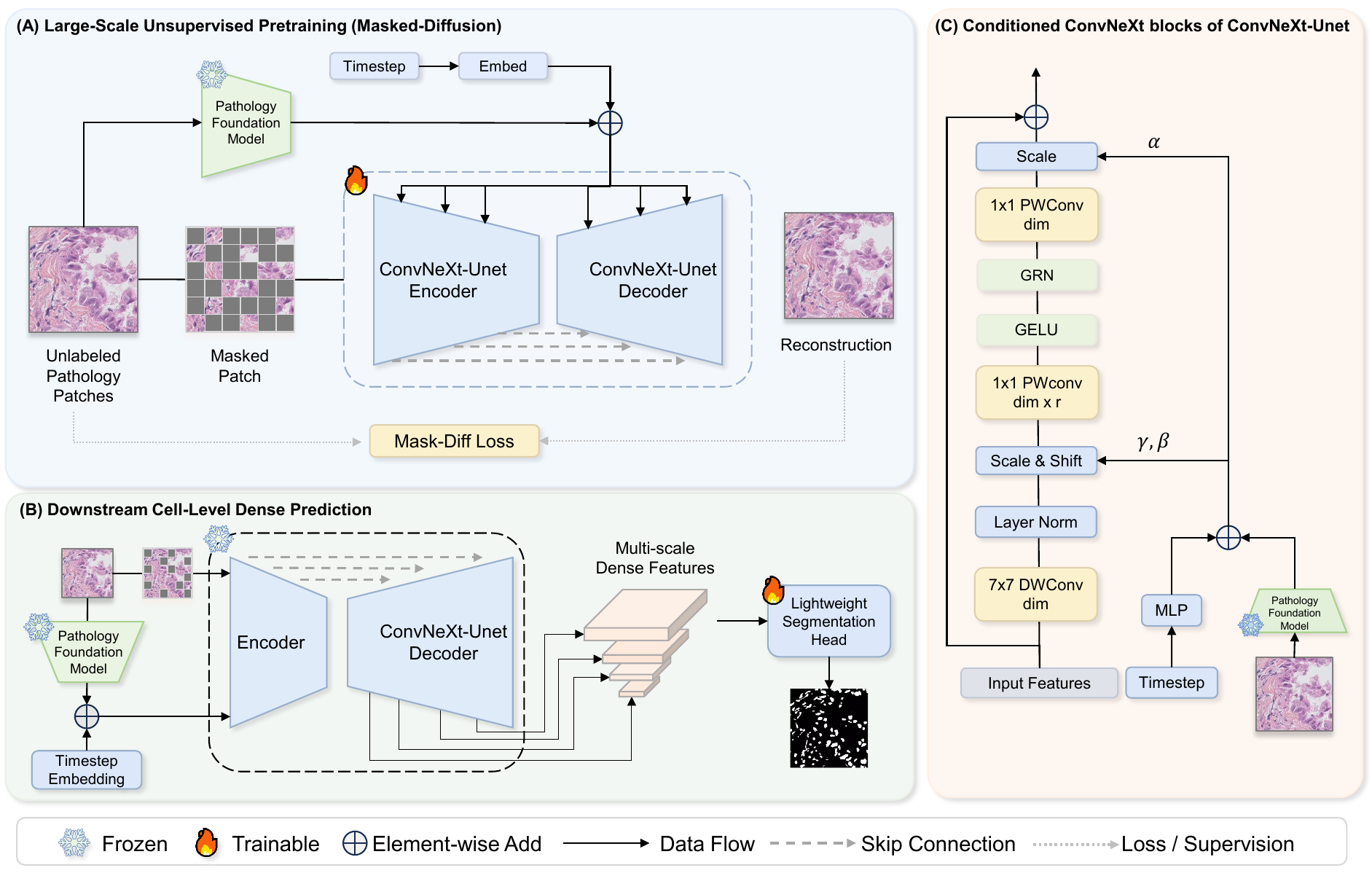}
    \caption{
    Overview of the proposed CMD framework. (A) Large-scale unlabeled pathology patches are used for self-supervised masked-diffusion pretraining, where timestep embeddings and pathology foundation model features condition a ConvNeXt U-Net to reconstruct masked patches. (B) After pretraining, the frozen generative backbone is reused as a dense visual encoder: multi-scale decoder features are extracted and fed into lightweight task-specific heads for cell-level dense prediction. (C) Each conditioned ConvNeXt block injects the fused timestep and pathology features via adaLN.
    }
    \label{fig:overview}
\end{figure}

This protocol separates representation learning from task-specific supervision: expensive pretraining is performed once on unlabeled pathology data, and downstream adaptation requires only limited annotations. It also allows us to evaluate whether masked-diffusion pretraining learns generalizable pathology representations, rather than relying on end-to-end supervised tuning. The overall pipeline is shown in Figure~\ref{fig:overview}.

\subsection{Design Question I: Why ConvNeXt-UNet as the Diffusion Backbone?}
\textbf{ConvNeXt-UNet Backbone for Microscopic Locality and Multi-Scale Structure}

The diffusion backbone determines how effectively masked-diffusion pretraining captures pathology morphology. For cell-level dense prediction, features must preserve nuclear contours, chromatin texture, thin boundaries, and multi-scale tissue context. We therefore compare DiT~\cite{peebles2023scalable}, Attention U-Net~\cite{oktay2018attention}, and ConvNeXt-UNet under the same pretraining and evaluation protocol.

DiT offers scalable global modeling, but patch tokenization may weaken small structures and intra-patch boundaries. Attention U-Net is naturally suited to dense prediction, yet its conventional blocks may limit scalability. ConvNeXt-UNet combines U-Net-style multi-scale feature reuse with modern ConvNeXt blocks, providing locality bias, efficient channel mixing, and adaptive conditioning with timestep and pathology foundation model features.

As shown in Table~\ref{tab:backbones}, ConvNeXt-UNet achieves stronger boundary-sensitive dense representations across datasets, supporting its use as the CMD diffusion backbone.

\begin{table}[htbp]
\centering
\caption{Backbone comparison under the same diffusion setting. DiT, Attention U-Net, and ConvNeXt-UNet all use VAE representations and pathology foundation model conditioning, are pretrained on 55K unlabeled pathology images with comparable parameter sizes, and are evaluated with frozen pretrained features and a linear-probe segmentation head. BF1 denotes boundary F1 score, measuring boundary agreement between predicted and ground-truth masks. Values in brackets denote 95\% confidence intervals estimated with 1000 bootstrap resamples. Detailed experimental settings are provided in Appendix~\ref{app:method_config}.}
\label{tab:backbones}
\resizebox{\textwidth}{!}{%
\small
\begin{tabular}{l *{5}{cc}}
\toprule
\multirow{2}{*}{\textbf{Backbone(linear-probe)}} & \multicolumn{2}{c}{\textbf{CPM-15~\cite{vu2019methods}}} & \multicolumn{2}{c}{\textbf{CPM-17~\cite{vu2019methods}}} & \multicolumn{2}{c}{\textbf{Janowczyk~\cite{janowczyk2016deep}}} & \multicolumn{2}{c}{\textbf{Kumar~\cite{graham2020dense}}} & \multicolumn{2}{c}{\textbf{TNBC~\cite{naylor2018segmentation}}} \\
\cmidrule(lr){2-3} \cmidrule(lr){4-5} \cmidrule(lr){6-7} \cmidrule(lr){8-9} \cmidrule(lr){10-11}
 & Dice & BF1 & Dice & BF1 & Dice & BF1 & Dice & BF1 & Dice & BF1 \\
\midrule
\shortstack[c]{\textbf{DiT} \\ \phantom{\tiny [0.00, 0.00]}} & \vc{44.39}{38.74}{50.04} & \vc{19.65}{17.71}{21.58} & \vc{56.33}{51.37}{61.12} & \vc{23.24}{12.60}{32.62} & \vc{50.56}{49.95}{51.26} & \vc{5.15}{3.36}{7.34} & \vc{48.80}{46.73}{52.28} & \vc{11.12}{9.45}{12.65} & \vc{55.11}{53.19}{56.97} & \vc{21.12}{15.63}{25.75} \\
\shortstack[c]{\textbf{Attention U-Net} \\ \phantom{\tiny [0.00, 0.00]}} & \vc{53.68}{49.98}{57.38} & \vc{19.31}{12.57}{26.04} & \vc{56.04}{53.71}{59.12} & \vc{23.19}{19.62}{26.25} & \vc{49.50}{49.24}{49.76} & \vc{1.67}{0.89}{2.59} & \vc{50.37}{47.53}{55.40} & \vc{20.80}{14.66}{28.45} & \vc{54.05}{51.91}{56.02} & \vc{16.39}{8.68}{22.57} \\
\shortstack[c]{\textbf{ConvNeXt-UNet} \\ \phantom{\tiny [0.00, 0.00]}} & \vc{50.19}{49.39}{50.99} & \vc{33.66}{31.42}{35.89} & \vc{55.20}{52.82}{57.39} & \vc{35.04}{27.83}{41.36} & \vc{50.27}{49.86}{50.75} & \vc{6.45}{4.24}{8.92} & \vc{51.61}{51.27}{51.90} & \vc{34.38}{31.67}{37.95} & \vc{53.23}{51.86}{54.60} & \vc{25.09}{14.05}{32.77} \\
\bottomrule
\end{tabular}%
}
\end{table}

\subsection{Design Question II: Why Masked-Diffusion Instead of Standard DDPM?}
\textbf{Masked-Diffusion Objective for Morphology-Aware Self-Supervision}

Inspired by masked diffusion~\cite{pan2023masked}, we treat diffusion pretraining as time-conditioned reconstruction rather than only generative sampling. From this view, the timestep controls corruption difficulty, and Gaussian noise in standard DDPM~\cite{ho2020denoising} can be replaced by a corruption process better aligned with representation learning.

For cell-level pathology, we use structure-oriented masking instead of Gaussian corruption. CMD learns to recover missing histological regions from visible context, encouraging representations that capture local morphology, tissue texture, nuclear boundaries, and spatial organization.

Given an unlabeled pathology patch $x_0$, we sample $t \in [1,T]$ and set $r_t=t/(T+1)$. Random non-overlapping patches are masked according to $r_t$ to obtain $x_t$, and the ConvNeXt-UNet reconstructs the original image or latent representation conditioned on the timestep and pathology foundation model feature:
\begin{align}
x_t &= \mathcal{M}(x_0, r_t), \quad r_t = \frac{t}{T+1}, \\
\hat{x}_0 &= f_\theta(x_t, t, z_\text{pfm}), \\
\mathcal{L}_\text{masked-diff} &=
\mathbb{E}_{x_0,t}
\left[
\left\| x_0 - f_\theta(x_t, t, z_\text{pfm}) \right\|_1
\right],
\end{align}
where $\mathcal{M}(\cdot)$ denotes timestep-controlled patch masking, $z_\text{pfm}$ is the frozen pathology foundation model feature, and $f_\theta$ is the trainable ConvNeXt-UNet. A detailed theoretical derivation is provided in Appendix~\ref{app:cmd_theory}.

Thus, $t$ changes from controlling Gaussian noise strength in DDPM to controlling structural occlusion in CMD. Timestep and pathology features are fused and injected into ConvNeXt blocks through adaptive Layer Normalization.

\subsection{Design Question III: Pixel-Space or VAE Latent-Space Pretraining?}
\textbf{Pixel-Space Masked Diffusion for Preserving Cell-Level Details}

Masked diffusion can operate in pixel space or compressed VAE~\cite{kingma2013auto} latent space. While latent-space diffusion is efficient, VAE compression may discard high-frequency cues such as nuclear boundaries, chromatin texture, and small inter-cell gaps, which are critical for pathology dense prediction.

We compare pixel-space pretraining, VAE latent-space pretraining, and a high-resolution VAE variant. As shown in Table~\ref{tab:pretraining_space}, pixel-space masked diffusion yields stronger downstream dense features, suggesting that preserving native pixel-level morphology is more important than latent-space efficiency for cell-level prediction.

\begin{table}[htbp]
\centering
\caption{Pixel-space versus VAE latent-space masked-diffusion pretraining. Pixel-space and standard VAE latent-space settings use $256 \times 256$ inputs, while the High-res VAE latent setting uses $512 \times 512$ inputs. Other experimental settings are the same as in Table~\ref{tab:backbones}.}
\label{tab:pretraining_space}
\resizebox{\textwidth}{!}{%
\small
\begin{tabular}{l *{5}{cc}}
\toprule
\multirow{2}{*}{\textbf{Pretraining Space}} & \multicolumn{2}{c}{\textbf{CPM-15}} & \multicolumn{2}{c}{\textbf{CPM-17}} & \multicolumn{2}{c}{\textbf{Janowczyk}} & \multicolumn{2}{c}{\textbf{Kumar}} & \multicolumn{2}{c}{\textbf{TNBC}} \\
\cmidrule(lr){2-3} \cmidrule(lr){4-5} \cmidrule(lr){6-7} \cmidrule(lr){8-9} \cmidrule(lr){10-11}
 & Dice & BF1 & Dice & BF1 & Dice & BF1 & Dice & BF1 & Dice & BF1 \\
\midrule
\shortstack[c]{\textbf{VAE latent-space} \\ \phantom{\tiny [0.00, 0.00]}} & \vc{50.19}{49.39}{50.99} & \vc{33.66}{31.42}{35.89} & \vc{55.20}{52.82}{57.39} & \vc{35.04}{27.83}{41.36} & \vc{50.27}{49.86}{50.75} & \vc{6.45}{4.24}{8.92} & \vc{51.61}{51.27}{51.90} & \vc{34.38}{31.67}{37.95} & \vc{53.23}{51.86}{54.60} & \vc{25.09}{14.05}{32.77} \\
\shortstack[c]{\textbf{High-res VAE latent} \\ \phantom{\tiny [0.00, 0.00]}} & \vc{47.96}{47.95}{47.98} & \vc{11.52}{9.54}{13.50} & \vc{63.80}{58.88}{68.49} & \vc{30.49}{24.61}{35.82} & \vc{51.87}{51.24}{52.63} & \vc{11.03}{9.00}{13.37} & \vc{53.09}{51.39}{54.40} & \vc{33.31}{30.64}{36.10} & \vc{58.24}{56.51}{59.81} & \vc{22.43}{19.24}{24.80} \\
\shortstack[c]{\textbf{Pixel-space} \\ \phantom{\tiny [0.00, 0.00]}} & \vc{63.25}{59.34}{67.16} & \vc{56.03}{55.41}{56.64} & \vc{76.74}{70.29}{82.39} & \vc{73.45}{62.26}{83.70} & \vc{53.66}{52.71}{54.70} & \vc{18.24}{13.99}{22.46} & \vc{62.76}{49.07}{70.45} & \vc{62.25}{26.32}{83.66} & \vc{73.51}{68.28}{80.45} & \vc{75.63}{66.74}{86.52} \\
\bottomrule
\end{tabular}%
}
\end{table}

\subsection{Design Question IV: Is Pathology Foundation Model Conditioning Essential?}
\textbf{Pathology Foundation Model Conditioning as Complementary Semantic Guidance}

Pathology foundation models provide global tissue context and high-level morphological semantics. CMD uses frozen pathology foundation model features as conditional guidance, complementing local masked reconstruction with pathology-aware semantic priors.

A frozen foundation model extracts an image-level feature from the original patch, which is fused with the timestep embedding and injected into ConvNeXt-UNet blocks through adaptive Layer Normalization. The dense representation is still learned by the masked-diffusion backbone.

We compare ConvNeXt masked diffusion without conditioning and full CMD with conditioning, using UNI as the foundation model. As shown in Table~\ref{tab:pfm_conditioning}, conditioning further improves performance, indicating that PFM features provide complementary semantic guidance rather than replacing masked-diffusion representation learning.

\begin{table}[htbp]
\centering
\caption{Effect of pathology foundation model conditioning. Other experimental settings are the same as in Table~\ref{tab:backbones}.}
\label{tab:pfm_conditioning}
\resizebox{\textwidth}{!}{%
\small
\begin{tabular}{l *{5}{cc}}
\toprule
\multirow{2}{*}{\textbf{Method}} & \multicolumn{2}{c}{\textbf{CPM-15}} & \multicolumn{2}{c}{\textbf{CPM-17}} & \multicolumn{2}{c}{\textbf{Janowczyk}} & \multicolumn{2}{c}{\textbf{Kumar}} & \multicolumn{2}{c}{\textbf{TNBC}} \\
\cmidrule(lr){2-3} \cmidrule(lr){4-5} \cmidrule(lr){6-7} \cmidrule(lr){8-9} \cmidrule(lr){10-11}
 & Dice & BF1 & Dice & BF1 & Dice & BF1 & Dice & BF1 & Dice & BF1 \\
\midrule
\shortstack[c]{\textbf{CMD w/o PFM} \\ \phantom{\tiny [0.00, 0.00]}} & \vc{55.17}{50.90}{59.45} & \vc{43.52}{40.68}{46.36} & \vc{71.29}{64.18}{78.05} & \vc{64.36}{50.59}{77.59} & \vc{52.21}{51.47}{53.08} & \vc{14.19}{10.53}{17.69} & \vc{60.21}{55.20}{66.46} & \vc{68.02}{61.82}{76.16} & \vc{60.24}{52.82}{67.65} & \vc{45.90}{32.53}{58.28} \\
\shortstack[c]{\textbf{CMD w/ PFM} \\ \phantom{\tiny [0.00, 0.00]}} & \vc{63.25}{59.34}{67.16} & \vc{56.03}{55.41}{56.64} & \vc{76.74}{70.29}{82.39} & \vc{73.45}{62.26}{83.70} & \vc{53.66}{52.71}{54.70} & \vc{18.24}{13.99}{22.46} & \vc{62.76}{49.07}{70.45} & \vc{62.25}{26.32}{83.66} & \vc{73.51}{68.28}{80.45} & \vc{75.63}{66.74}{86.52} \\
\bottomrule
\end{tabular}%
}
\end{table}

\section{Experiments}
\label{others}

\subsection{Experimental Setup}

\subsubsection{Pretraining data}

We pretrain CMD on a large-scale unlabeled pathology corpus with approximately 1 million $512 \times 512$ image patches. All patches are unlabeled and used only for self-supervised masked-diffusion pretraining. Additional dataset details are provided in Appendix~\ref{app:datasets}.

\subsubsection{Pretraining protocol}

During pretraining, each $512 \times 512$ pathology patch is converted into a $256 \times 256$ input. We use a mixed resizing strategy: with 80\% probability, we randomly crop a $256 \times 256$ region, and with 20\% probability, we resize the whole patch to $256 \times 256$. This exposes the model to both fine cellular details and broader tissue layouts. 

We use masked diffusion with $T=1000$ timesteps. For a sampled timestep $t$, the masking ratio is $r_t=t/T$, and random non-overlapping patches are masked and reconstructed by the diffusion backbone. For pixel-space pretraining, the mask patch size is 8, yielding a $32 \times 32$ patch grid for $256 \times 256$ inputs. When pathology foundation model conditioning is enabled, we use a frozen H0-mini encoder and inject its image-level feature together with the timestep embedding into ConvNeXt blocks through adaptive Layer Normalization.

The model is optimized with AdamW using a learning rate of $3 \times 10^{-5}$, no weight decay, BF16 mixed precision, and an exponential moving average decay of 0.9999. Unless otherwise specified, we train for 80K optimization steps and use the EMA checkpoint for downstream evaluation.

\subsubsection{Downstream task protocol}

For downstream cell-level dense prediction, we freeze the pretrained CMD  backbone and use it only as a dense feature extractor. Multi-scale features are extracted from selected decoder blocks and passed to a task-specific segmentation head. During downstream training, only the segmentation head is optimized, using a cosine learning-rate schedule and a combined cross-entropy plus Dice loss. Details of the downstream head designs are provided in Appendix~\ref{app:downstream_head}.

\subsection{Comparisons With State-of-the-art Methods}

We evaluate CMD on cell-level dense prediction under the frozen-backbone setting and compare it with both frozen pathology foundation models and end-to-end segmentation baselines. The frozen-backbone comparison isolates representation quality, while the end-to-end baselines provide strong task-specific references trained directly for segmentation.

As shown in Table~\ref{tab:model_perf_param_frozen_pfm}, CMD achieves stronger overall performance than ViT-based pathology foundation models and remains competitive with, or superior to, end-to-end segmentation models across CPM-15, CPM-17, and TNBC. This indicates that CMD is not only a stronger frozen pathology representation, but also provides dense features that can rival specialized segmentation architectures.

An important advantage of CMD is its reduced sensitivity to input resolution. CMD uses a unified $256 \times 256$ downstream input, whereas ViT-based foundation models and end-to-end baselines often require larger dataset-specific resolutions. Despite this smaller input size, CMD matches or outperforms high-resolution ViT-based models, suggesting that the convolutional masked-diffusion representation captures local morphology efficiently without relying on larger image fields.

\begin{figure}[h]
\centering
\includegraphics[width=\linewidth]{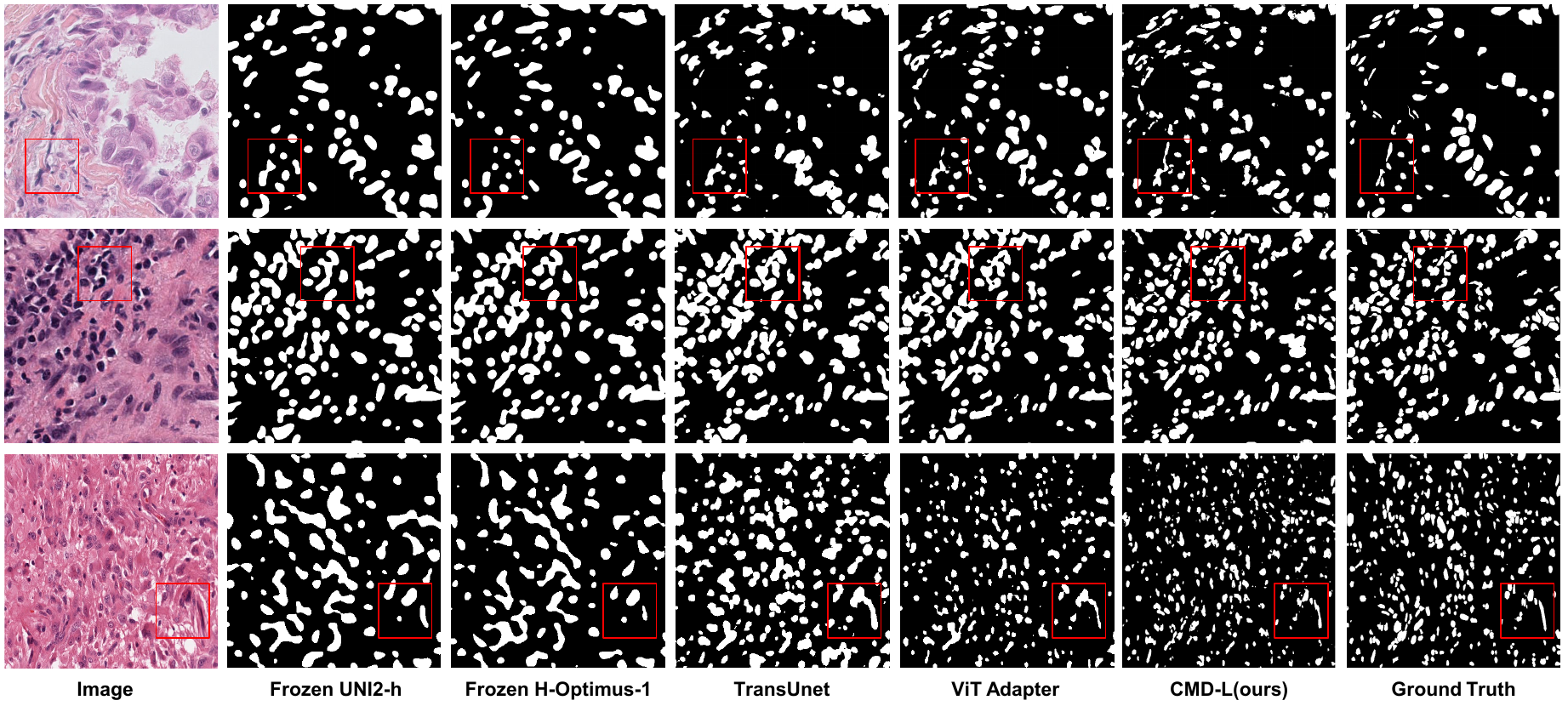}
\caption{Qualitative comparison under the frozen-backbone dense prediction setting. CMD-L produces more complete and boundary-consistent cell masks than frozen ViT-based pathology foundation models and supervised dense prediction baselines, especially in crowded or ambiguous regions highlighted by red boxes. From top to bottom are the TNBC, CPM17, and CPM15 datasets.}
\label{fig:qualitative_frozen_backbone}
\end{figure}

Figure~\ref{fig:qualitative_frozen_backbone} further supports this observation. Frozen ViT-based foundation models often capture coarse foreground regions but miss small nuclei, merge adjacent cells, or produce noisy boundaries in crowded tissue areas. End-to-end baselines improve spatial layout but can still produce fragmented boundaries or incomplete cell separation. In contrast, CMD-L better preserves small-cell structures, boundary consistency, and separation between adjacent nuclei, especially in the highlighted challenging regions.

Overall, these results show that CMD is an effective frozen backbone for pathology dense prediction. Its convolutional architecture and masked-diffusion pretraining make it less dependent on high input resolution while maintaining strong morphology-aware segmentation performance.

\begin{table}[htbp]
\centering
\caption{Segmentation performance and parameter statistics on CPM-15, CPM-17, and TNBC. The table compares frozen pathology foundation model backbones, end-to-end segmentation baselines, and CMD. Dataset headers report the input resolutions used by different model families. Results are reported as mean Dice/Precision with 95\% confidence intervals.}
\label{tab:model_perf_param_frozen_pfm}
\resizebox{\textwidth}{!}{%
\small
\begin{tabular}{l cc *{3}{cc}}
\toprule
\multirow{2}{*}{\textbf{Method}} & \multirow{2}{*}{\shortstack[c]{\textbf{Frozen}\\\textbf{Params. (M)}}} & \multirow{2}{*}{\shortstack[c]{\textbf{Seg. Head}\\\textbf{Params. (M)}}} & \multicolumn{2}{c}{\shortstack[c]{\textbf{cpm15}\\\footnotesize ViT/16:384; ViT/14:392; CMD:256\\\footnotesize E2E Model:384}} & \multicolumn{2}{c}{\shortstack[c]{\textbf{cpm17}\\\footnotesize ViT/16:496; ViT/14:490; CMD:256\\\footnotesize E2E Model:496}} & \multicolumn{2}{c}{\shortstack[c]{\textbf{TNBC}\\\footnotesize ViT/16:512; ViT/14:504; CMD:256\\\footnotesize E2E Model:512}} \\
\cmidrule(lr){4-5} \cmidrule(lr){6-7} \cmidrule(lr){8-9}
& & & Dice & Precision & Dice & Precision & Dice & Precision \\
\midrule
\multicolumn{9}{c}{\textbf{Pre-trained ViT-based Pathology Foundation Models}} \\
\midrule
\textbf{PathOrchestra} & 303 & 7.06 & \vc{0.737}{0.658}{0.822} & \vc{0.758}{0.681}{0.838} & \vc{0.861}{0.837}{0.882} & \vc{0.856}{0.827}{0.888} & \vc{0.819}{0.781}{0.855} & \vc{0.810}{0.764}{0.850} \\
\textbf{CONCHv1.5} & 306 & 7.06 & \vc{0.748}{0.670}{0.824} & \vc{0.740}{0.690}{0.787} & \vc{0.858}{0.829}{0.885} & \vc{0.863}{0.834}{0.900} & \vc{0.826}{0.793}{0.859} & \vc{0.806}{0.763}{0.845} \\
\textbf{CONCH} & 90.4 & 5.88 & \vc{0.722}{0.608}{0.825} & \vc{0.746}{0.689}{0.793} & \vc{0.867}{0.841}{0.889} & \vc{0.877}{0.844}{0.906} & \vc{0.816}{0.785}{0.847} & \vc{0.793}{0.771}{0.822} \\
\textbf{Prov-Gigapath} & 1100 & 9.42 & \vc{0.759}{0.680}{0.842} & \vc{0.753}{0.685}{0.822} & \vc{0.871}{0.851}{0.888} & \vc{0.863}{0.832}{0.893} & \vc{0.828}{0.795}{0.860} & \vc{0.809}{0.773}{0.842} \\
\textbf{Hibou} & 304 & 7.06 & \vc{0.794}{0.722}{0.874} & \vc{0.781}{0.715}{0.853} & \vc{0.875}{0.853}{0.896} & \vc{0.872}{0.837}{0.915} & \vc{0.841}{0.812}{0.866} & \vc{0.823}{0.793}{0.855} \\
\textbf{H-Optimus-0} & 1100 & 9.42 & \vc{0.781}{0.700}{0.868} & \vc{0.795}{0.721}{0.871} & \vc{0.881}{0.862}{0.899} & \vc{0.869}{0.839}{0.904} & \vc{0.843}{0.808}{0.869} & \vc{0.811}{0.773}{0.842} \\
\textbf{H-Optimus-1} & 1100 & 9.42 & \vc{0.773}{0.711}{0.846} & \vc{0.771}{0.696}{0.871} & \vc{0.879}{0.857}{0.896} & \vc{0.870}{0.841}{0.895} & \vc{0.833}{0.805}{0.859} & \vc{0.802}{0.775}{0.827} \\
\textbf{Kaiko} & 304 & 7.06 & \vc{0.758}{0.663}{0.855} & \vc{0.803}{0.733}{0.866} & \vc{0.882}{0.861}{0.899} & \vc{0.866}{0.839}{0.894} & \vc{0.838}{0.807}{0.867} & \vc{0.822}{0.790}{0.852} \\
\textbf{Midnight-12k} & 1100 & 9.42 & \vc{0.710}{0.650}{0.775} & \vc{0.715}{0.660}{0.772} & \vc{0.845}{0.821}{0.865} & \vc{0.831}{0.806}{0.854} & \vc{0.798}{0.761}{0.834} & \vc{0.771}{0.731}{0.809} \\
\textbf{MUSK} & 675 & 7.06 & \vc{0.651}{0.614}{0.693} & \vc{0.643}{0.608}{0.681} & \vc{0.737}{0.706}{0.766} & \vc{0.728}{0.696}{0.762} & \vc{0.693}{0.666}{0.720} & \vc{0.668}{0.642}{0.696} \\
\textbf{Phikon} & 86.4 & 5.88 & \vc{0.759}{0.686}{0.837} & \vc{0.752}{0.690}{0.815} & \vc{0.871}{0.844}{0.897} & \vc{0.873}{0.833}{0.906} & \vc{0.828}{0.800}{0.856} & \vc{0.804}{0.776}{0.840} \\
\textbf{Phikon-v2} & 303 & 7.06 & \vc{0.754}{0.678}{0.841} & \vc{0.750}{0.670}{0.846} & \vc{0.874}{0.854}{0.893} & \vc{0.868}{0.842}{0.895} & \vc{0.824}{0.800}{0.849} & \vc{0.794}{0.777}{0.809} \\
\textbf{UNI} & 303 & 7.06 & \vc{0.738}{0.662}{0.813} & \vc{0.726}{0.674}{0.775} & \vc{0.867}{0.838}{0.892} & \vc{0.854}{0.812}{0.890} & \vc{0.820}{0.790}{0.850} & \vc{0.786}{0.756}{0.811} \\
\textbf{UNI2-h} & 681 & 9.42 & \vc{0.767}{0.703}{0.840} & \vc{0.761}{0.696}{0.835} & \vc{0.873}{0.852}{0.890} & \vc{0.864}{0.835}{0.891} & \vc{0.833}{0.807}{0.855} & \vc{0.804}{0.775}{0.829} \\
\textbf{Virchow} & 631 & 8.24 & \vc{0.770}{0.701}{0.853} & \vc{0.767}{0.685}{0.880} & \vc{0.876}{0.859}{0.890} & \vc{0.873}{0.853}{0.897} & \vc{0.839}{0.817}{0.861} & \vc{0.813}{0.789}{0.837} \\
\textbf{Virchow2} & 631 & 8.24 & \vc{0.725}{0.643}{0.827} & \vc{0.707}{0.640}{0.817} & \vc{0.863}{0.832}{0.887} & \vc{0.851}{0.811}{0.888} & \vc{0.821}{0.801}{0.841} & \vc{0.778}{0.759}{0.796} \\
\midrule
\multicolumn{9}{c}{\textbf{End-to-End Segmentation Models}} \\
\midrule
\textbf{nnUNet} & - & \shortstack[c]{12.19} & \vc{0.791}{0.765}{0.820} & \vc{0.751}{0.732}{0.772} & \vc{0.875}{0.859}{0.898} & \vc{0.842}{0.819}{0.874} & \vc{0.803}{0.778}{0.822} & \vc{0.754}{0.719}{0.779} \\
\textbf{TransUNet} & - & \shortstack[c]{105.28} & \vc{0.809}{0.771}{0.850} & \vc{0.770}{0.739}{0.803} & \vc{0.897}{0.882}{0.914} & \vc{0.881}{0.856}{0.909} & \vc{0.843}{0.822}{0.860} & \vc{0.825}{0.794}{0.853} \\
\textbf{ViT-Adapter} & - & \shortstack[c]{133.52} & \vc{0.876}{0.830}{0.924} & \vc{0.889}{0.863}{0.916} & \vc{0.872}{0.839}{0.898} & \vc{0.877}{0.838}{0.917} & \vc{0.854}{0.832}{0.876} & \vc{0.838}{0.805}{0.872} \\
\midrule
\multicolumn{9}{c}{\textbf{SAM Pre-trained ViT-based Pathology Foundation Models}} \\
\midrule
\textbf{VISTA~\cite{liang2026vista}} & 168 & Zero-shot & \vc{0.208}{0.149}{0.266} & \vc{0.518}{0.487}{0.576} & \vc{0.481}{0.366}{0.560} & \vc{0.531}{0.489}{0.570} & \vc{0.551}{0.501}{0.607} & \vc{0.584}{0.562}{0.607} \\
\midrule
\multicolumn{9}{c}{\color{darkgreen}\textbf{Pre-trained Purely Convolutional Pathology Foundation Models}} \\
\midrule
\textbf{CMD-L(ours)} & 602 & Linear Probe & \vc{0.676}{0.643}{0.709} & \vc{0.696}{0.636}{0.756} & \vc{0.841}{0.811}{0.871} & \vc{0.878}{0.848}{0.901} & \vc{0.807}{0.774}{0.841} & \vc{0.813}{0.783}{0.836} \\
\shortstack[l]{\textbf{CMD-L(ours)}\\\footnotesize  + SegHead: Fig.~\ref{fig:downstream_heads}(A)} & 602 & 7.15 & \vc{0.864}{0.845}{0.882} & \vc{0.849}{0.842}{0.857} & \vc{0.865}{0.827}{0.902} & \vc{0.901}{0.876}{0.925} & \vc{0.854}{0.824}{0.891} & \vc{0.851}{0.822}{0.876} \\
\shortstack[l]{\textbf{CMD-L(ours)}\\\footnotesize  + SegHead: Fig.~\ref{fig:downstream_heads}(B)} & 602 & 21.21 & \textbf{\vc{0.889}{0.859}{0.925}} & \textbf{\vc{0.885}{0.853}{0.925}} & \textbf{\vc{0.880}{0.847}{0.908}} & \textbf{\vc{0.883}{0.860}{0.920}} & \textbf{\vc{0.854}{0.835}{0.874}} & \textbf{\vc{0.830}{0.802}{0.858}} \\
\bottomrule
\end{tabular}%
}
\end{table}

\subsection{State-of-the-Art Parameter-Efficient Few-Shot Adaptation}

We further evaluate few-shot transfer for cell-level dense prediction on CPM-17 and TNBC under 1-shot, 5-shot, and 10-shot settings. This setting tests whether pretrained representations can be adapted with very limited annotations. We compare CMD-L with frozen pathology foundation model baselines, TransUNet with frozen UNI2-h features, and an end-to-end ViT-Adapter baseline. For CMD-L, the pretrained ConvNeXt masked-diffusion backbone remains frozen, and only a segmentation head with 7M parameters ~\cite{li2023open} is optimized.

As shown in Table~\ref{tab:fewshot_performance}, frozen pathology foundation models provide limited few-shot segmentation performance, indicating that global pathology representations are not sufficient for dense cell-level prediction. TransUNet improves slightly in some settings, but still struggles to recover fine-grained nuclei structures from frozen image-level features. ViT-Adapter achieves strong results when more labels are available, but requires end-to-end optimization of a much larger number of trainable parameters.

In contrast, CMD-L achieves competitive or superior performance with only 7M trainable parameters. It performs especially well in the most label-scarce settings and maintains strong Dice and Precision as the number of shots increases. These results show that CMD-L learns morphology-aware dense features that are both label-efficient and parameter-efficient for few-shot cell-level segmentation.

\begin{table}[htbp]
\centering
\caption{Few-shot segmentation performance and parameter statistics on CPM-17 and TNBC. The table compares frozen pathology foundation models, TransUNet, end-to-end ViT-Adapter, and CMD-L under 1-shot, 5-shot, and 10-shot settings. Results are reported as mean Dice/Precision with 95\% confidence intervals.}
\label{tab:fewshot_performance}
\resizebox{\textwidth}{!}{%
\small
\begin{tabular}{l *{5}{cc}}
\toprule
\multirow{2}{*}{\textbf{Condition}} & \multicolumn{2}{c}{\textbf{Frozen UNI2-h}} & \multicolumn{2}{c}{\textbf{Frozen H-opt-1}} & \multicolumn{2}{c}{\textbf{TransUNet}} & \multicolumn{2}{c}{\textbf{ViT-Adapter}} & \multicolumn{2}{c}{\textbf{Frozen CMD-L(ours)}} \\
\cmidrule(lr){2-3} \cmidrule(lr){4-5} \cmidrule(lr){6-7} \cmidrule(lr){8-9} \cmidrule(lr){10-11}
& Dice & Precision & Dice & Precision & Dice & Precision & Dice & Precision & Dice & Precision \\
\midrule
\textbf{Trainable/Total Params. (M)} & \multicolumn{2}{c}{9/690} & \multicolumn{2}{c}{8/1144} & \multicolumn{2}{c}{23/704} & \multicolumn{2}{c}{364/364} & \multicolumn{2}{c}{7/609} \\
\midrule
\textbf{cpm17 - 1-Shot} & \vc{0.463}{0.454}{0.472} & \vc{0.507}{0.490}{0.528} & \vc{0.468}{0.458}{0.479} & \vc{0.493}{0.485}{0.502} & \vc{0.479}{0.464}{0.492} & \vc{0.499}{0.485}{0.514} & \vc{0.714}{0.683}{0.749} & \vc{0.695}{0.661}{0.740} & \vc{0.769}{0.696}{0.832} & \vc{0.784}{0.718}{0.845} \\
\textbf{cpm17 - 5-Shot} & \vc{0.468}{0.457}{0.477} & \vc{0.501}{0.487}{0.521} & \vc{0.467}{0.456}{0.480} & \vc{0.497}{0.483}{0.512} & \vc{0.477}{0.463}{0.489} & \vc{0.499}{0.487}{0.511} & \vc{0.812}{0.694}{0.880} & \vc{0.852}{0.806}{0.904} & \vc{0.811}{0.735}{0.882} & \vc{0.893}{0.863}{0.918} \\
\textbf{cpm17 - 10-Shot} & \vc{0.464}{0.453}{0.474} & \vc{0.500}{0.487}{0.518} & \vc{0.472}{0.460}{0.485} & \vc{0.508}{0.489}{0.529} & \vc{0.476}{0.463}{0.488} & \vc{0.497}{0.485}{0.509} & \vc{0.819}{0.725}{0.881} & \vc{0.832}{0.780}{0.894} & \vc{0.820}{0.744}{0.887} & \vc{0.896}{0.875}{0.914} \\
\midrule
\textbf{TNBC - 1-Shot} & \vc{0.497}{0.477}{0.519} & \vc{0.506}{0.473}{0.534} & \vc{0.486}{0.477}{0.497} & \vc{0.500}{0.484}{0.516} & \vc{0.478}{0.469}{0.489} & \vc{0.495}{0.482}{0.512} & \vc{0.498}{0.470}{0.529} & \vc{0.501}{0.464}{0.539} & \vc{0.714}{0.618}{0.817} & \vc{0.865}{0.817}{0.917} \\
\textbf{TNBC - 5-Shot} & \vc{0.493}{0.482}{0.506} & \vc{0.509}{0.492}{0.529} & \vc{0.490}{0.480}{0.500} & \vc{0.508}{0.488}{0.529} & \vc{0.477}{0.468}{0.488} & \vc{0.484}{0.470}{0.500} & \vc{0.800}{0.773}{0.830} & \vc{0.808}{0.783}{0.849} & \vc{0.792}{0.761}{0.851} & \vc{0.845}{0.800}{0.880} \\
\textbf{TNBC - 10-Shot} & \vc{0.497}{0.491}{0.504} & \vc{0.514}{0.499}{0.525} & \vc{0.492}{0.484}{0.503} & \vc{0.510}{0.495}{0.524} & \vc{0.476}{0.468}{0.488} & \vc{0.910}{0.879}{0.953} & \vc{0.828}{0.807}{0.852} & \vc{0.837}{0.796}{0.916} & \vc{0.819}{0.790}{0.870} & \vc{0.845}{0.807}{0.877} \\
\bottomrule
\end{tabular}%
}
\end{table}

\subsection{Scaling Behavior in Pretraining Duration and Model Capacity}

We study CMD scaling from two aspects: masked-diffusion pretraining duration and diffusion backbone capacity. Unlike the few-shot setting, this analysis trains the downstream segmentation head with the full training split and evaluates on the corresponding test split, allowing us to assess representation quality under the standard full-data protocol.

Table~\ref{tab:training_steps_dice} shows that CMD generally benefits from longer pretraining. Performance improves or remains stable as the number of pretraining steps increases, indicating that masked-diffusion pretraining continues to produce transferable dense representations rather than overfitting to the reconstruction objective. Some dataset-specific fluctuations appear across intermediate checkpoints, but later checkpoints do not show systematic degradation.

We further compare CMD-B and CMD-L in Table~\ref{tab:fcdm_comparison}. Increasing backbone capacity consistently improves Dice across CPM-17 and TNBC, while maintaining comparable Precision. This suggests that larger ConvNeXt diffusion backbones improve region-level segmentation quality without increasing false positives. Architecture details of CMD-B and CMD-L are provided in Appendix~\ref{app:cmd_architecture}.

Overall, CMD shows favorable scaling behavior with both pretraining duration and model capacity. Although not a full scaling-law analysis, these results indicate that masked-diffusion convolutional pretraining can continue to benefit from more computation and larger backbones for cell-level dense prediction.

\begin{table}[htbp]
\centering
\caption{Effect of CMD-L pretraining duration on full-data downstream segmentation. Results are Dice scores reported as mean with 95\% confidence interval.}
\label{tab:training_steps_dice}
\resizebox{\textwidth}{!}{%
\small
\begin{tabular}{l *{8}{c}}
\toprule
\multirow{2}{*}{\textbf{Dataset}} & \multicolumn{8}{c}{\textbf{Training Steps}} \\
\cmidrule(lr){2-9}
& \textbf{10k} & \textbf{20k} & \textbf{30k} & \textbf{40k} & \textbf{50k} & \textbf{60k} & \textbf{70k} & \textbf{80k} \\
\midrule
\textbf{CPM15} & \vc{0.851}{0.806}{0.895} & \vc{0.858}{0.842}{0.874} & \vc{0.774}{0.751}{0.797} & \vc{0.859}{0.840}{0.879} & \vc{0.868}{0.845}{0.892} & \vc{0.866}{0.842}{0.889} & \vc{0.865}{0.841}{0.889} & \vc{0.864}{0.845}{0.882} \\
\textbf{CPM17} & \vc{0.873}{0.845}{0.903} & \vc{0.864}{0.829}{0.901} & \vc{0.827}{0.757}{0.899} & \vc{0.842}{0.787}{0.895} & \vc{0.851}{0.802}{0.899} & \vc{0.849}{0.796}{0.902} & \vc{0.863}{0.825}{0.902} & \vc{0.865}{0.827}{0.902} \\
\textbf{TNBC}  & \vc{0.820}{0.779}{0.884} & \vc{0.823}{0.775}{0.889} & \vc{0.829}{0.790}{0.877} & \vc{0.839}{0.802}{0.892} & \vc{0.844}{0.810}{0.888} & \vc{0.845}{0.814}{0.885} & \vc{0.850}{0.817}{0.892} & \vc{0.854}{0.824}{0.891} \\
\bottomrule
\end{tabular}%
}
\end{table}

\begin{figure}[t]
\centering
\includegraphics[width=\linewidth]{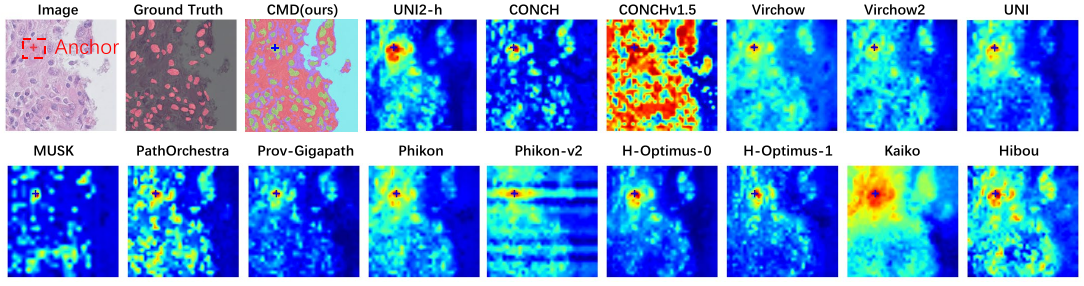}
\caption{Visualization of cell-level dense representations. For ViT-based pathology foundation models, we select an anchor patch on a nucleus and visualize cosine similarity between the anchor token and all other patch tokens. These models mainly highlight coarse tissue regions and show limited sensitivity to individual cellular structures. For CMD, which does not use patch tokens, we extract dense convolutional features and visualize their $K$-means ($K$=4) clustering result. CMD produces a more cell-aware representation that better follows fine nuclei distribution and local morphology.}
\label{fig:cell_level_representation_visualization}
\end{figure}

\begin{table}[htbp]
\centering
\caption{Backbone capacity scaling of CMD at 70k pretraining steps on full-data downstream segmentation. Results are mean with 95\% confidence interval.}
\label{tab:fcdm_comparison}
\footnotesize
\setlength{\tabcolsep}{3pt}
\begin{tabular}{l c c c c}
\toprule
\multirow{2}{*}{\textbf{Dataset}} & \multicolumn{2}{c}{\textbf{CMD-B}} & \multicolumn{2}{c}{\textbf{CMD-L}} \\
\cmidrule(lr){2-3} \cmidrule(lr){4-5}
& Dice & Precision & Dice & Precision \\
\midrule
\textbf{Params. (M) } & \multicolumn{2}{c}{217} & \multicolumn{2}{c}{602} \\
\midrule
\textbf{CPM15} & \vc{0.852}{0.836}{0.869} & \vc{0.843}{0.837}{0.849} & \vc{0.865}{0.841}{0.889} & \vc{0.852}{0.835}{0.869} \\
\textbf{CPM17} & \vc{0.829}{0.755}{0.902} & \vc{0.900}{0.870}{0.928} & \vc{0.863}{0.825}{0.902} & \vc{0.899}{0.872}{0.924} \\
\textbf{TNBC}  & \vc{0.836}{0.806}{0.875} & \vc{0.828}{0.797}{0.854} & \vc{0.850}{0.817}{0.892} & \vc{0.846}{0.812}{0.877} \\
\bottomrule
\end{tabular}
\end{table}

\subsection{Visualization of Cell-Level Dense Representations}

We visualize learned representations to examine whether CMD captures fine cellular structures rather than only coarse tissue semantics. As shown in Figure~\ref{fig:cell_level_representation_visualization}, ViT-based models tend to highlight broad tissue regions or coarse contextual patterns, with activations often spreading beyond individual nuclei. In contrast, CMD produces more localized and cell-aware feature clusters that better align with nuclei and fine boundaries. This supports the quantitative results: convolutional masked-diffusion pretraining learns dense, morphology-sensitive representations that are better suited for cell-level prediction.

\section{Conclusion and Future Work}

In this article, we introduced Masked-Diffusion Convolutional Foundation Models, termed \textbf{C}onvNeXt \textbf{M}asked-\textbf{D}iffusion (\textbf{CMD}), for cell-level dense prediction in computational pathology. CMD uses a fully convolutional ConvNeXt-UNet backbone with pixel-space masked-diffusion pretraining to learn dense, morphology-aware representations, while incorporating pathology foundation model conditioning as semantic guidance.

Experiments across multiple pathology benchmarks show that CMD provides strong frozen dense features, outperforming ViT-based pathology foundation models and remaining competitive with state-of-the-art end-to-end segmentation baselines. CMD also demonstrates parameter-efficient few-shot adaptation, favorable scaling behavior, and localized cell-aware representations in visualization analyses.

Future work will explore larger corpora and backbones, multimodal or report-guided conditioning, and whole-slide workflows linking cell-level prediction to slide-level diagnosis and tumor microenvironment analysis.


\bibliographystyle{unsrt} 
\begin{small}
\bibliography{neurips_2026_arxiv}
\end{small}
 
\newpage
\appendix

\section{Theoretical Overview of ConvNeXt Masked-Diffusion Models}
\label{app:cmd_theory}

\setcounter{equation}{0}

Masked-diffusion pretraining can be viewed as a self-supervised relaxation of denoising diffusion models for dense representation learning. In a standard DDPM, an image $x_0 \sim q(x_0)$ is corrupted by Gaussian noise at timestep $t$, and a time-conditioned network is trained to predict either the injected noise or the clean image. In contrast, masked diffusion replaces Gaussian corruption with timestep-controlled structural masking. This change removes the requirement that the forward process correspond to a valid generative diffusion chain, but preserves the key representation-learning property: the model must recover the original signal from corruptions of varying difficulty.

Given an unlabeled pathology image patch $x_0 \in \mathbb{R}^{H \times W \times C}$, we divide it into $N = HW/P^2$ non-overlapping patches with patch size $P$. A timestep $t \sim \mathcal{U}\{1,\ldots,T\}$ determines the masking ratio
\begin{equation}
    r_t = \frac{t}{T+1}.
\end{equation}
Let $m_t \in \{0,1\}^{H \times W \times C}$ be a binary patch mask obtained by randomly masking $\lfloor r_t N \rfloor$ patches and broadcasting the patch-level mask to pixels. The corrupted input is
\begin{equation}
    x_t = \mathcal{M}(x_0, t) = m_t \odot x_0,
\end{equation}
where $\odot$ denotes element-wise multiplication. Larger timesteps therefore correspond to stronger structural occlusion rather than stronger Gaussian noise. The model is trained to reconstruct the clean image from the partially observed input:
\begin{equation}
    \hat{x}_0 = f_\theta(x_t, t, z_{\mathrm{pfm}}),
\end{equation}
where $f_\theta$ is a ConvNeXt-U-Net masked-diffusion backbone and $z_{\mathrm{pfm}}$ denotes an optional frozen pathology foundation model feature.

The ConvNeXt-U-Net architecture provides a locality-preserving alternative to token-based diffusion backbones. Its encoder-decoder structure extracts multi-scale dense features, while skip connections preserve high-resolution morphology. Timestep and pathology conditions are projected into a shared conditioning vector,
\begin{equation}
    c = \phi_t(t) + \phi_z(z_{\mathrm{pfm}}),
\end{equation}
and injected into ConvNeXt blocks through adaptive normalization, e.g.,
\begin{equation}
    \operatorname{AdaLN}(u,c) = \gamma(c) \odot \operatorname{LN}(u) + \beta(c),
\end{equation}
where $\gamma(c)$ and $\beta(c)$ are learned scale and shift parameters. This allows the same convolutional backbone to adapt its reconstruction behavior to both the corruption level and pathology-aware semantic context.

Following the masked diffusion formulation, the pretraining loss compares the reconstruction $\hat{x}_0$ with the original image $x_0$. A pixel reconstruction objective can be written as
\begin{equation}
    \mathcal{L}_{\mathrm{rec}}
    =
    \mathbb{E}_{x_0,t}
    \left[
    \left\|x_0 - f_\theta(x_t,t,z_{\mathrm{pfm}})\right\|_1
    \right].
\end{equation}
To better align reconstruction pretraining with downstream dense prediction, one may also use a structural similarity loss:
\begin{equation}
    \mathcal{L}_{\mathrm{SSIM}}(x_0,\hat{x}_0)
    =
    \frac{1-\operatorname{SSIM}(x_0,\hat{x}_0)}{2},
\end{equation}
where
\begin{equation}
    \operatorname{SSIM}(x,\hat{x})
    =
    \frac{(2\mu_x\mu_{\hat{x}}+c_1)(2\sigma_{x\hat{x}}+c_2)}
    {(\mu_x^2+\mu_{\hat{x}}^2+c_1)(\sigma_x^2+\sigma_{\hat{x}}^2+c_2)}.
\end{equation}
Here $\mu$, $\sigma^2$, and $\sigma_{x\hat{x}}$ denote local means, variances, and covariance, while $c_1$ and $c_2$ stabilize the division. The overall masked-diffusion objective can therefore be expressed as
\begin{equation}
    \mathcal{L}_{\mathrm{CMD}}
    =
    \mathbb{E}_{x_0,t}
    \left[
    \lambda_1 \left\|x_0-\hat{x}_0\right\|_1
    +
    \lambda_s \mathcal{L}_{\mathrm{SSIM}}(x_0,\hat{x}_0)
    \right],
\end{equation}
with $\lambda_1$ and $\lambda_s$ controlling the balance between pixel fidelity and structural consistency.

After pretraining, the generative reconstruction head is not used for sampling. Instead, the frozen ConvNeXt masked-diffusion model serves as a dense feature extractor. Multi-scale decoder activations are collected at a fixed timestep, resized to a common resolution, and passed to a lightweight segmentation head. Thus, the model transfers the structural representations learned from unlabeled masked reconstruction to cell-level dense prediction with limited annotation.

\section{Experimental Details}
\renewcommand{\thefigure}{B\arabic{figure}}
\renewcommand{\thetable}{B\arabic{table}}
\setcounter{figure}{0}
\setcounter{table}{0}

\subsection{Datasets}
\label{app:datasets}

\begin{figure}[htbp!]
\centering
\includegraphics[width=0.85\linewidth]{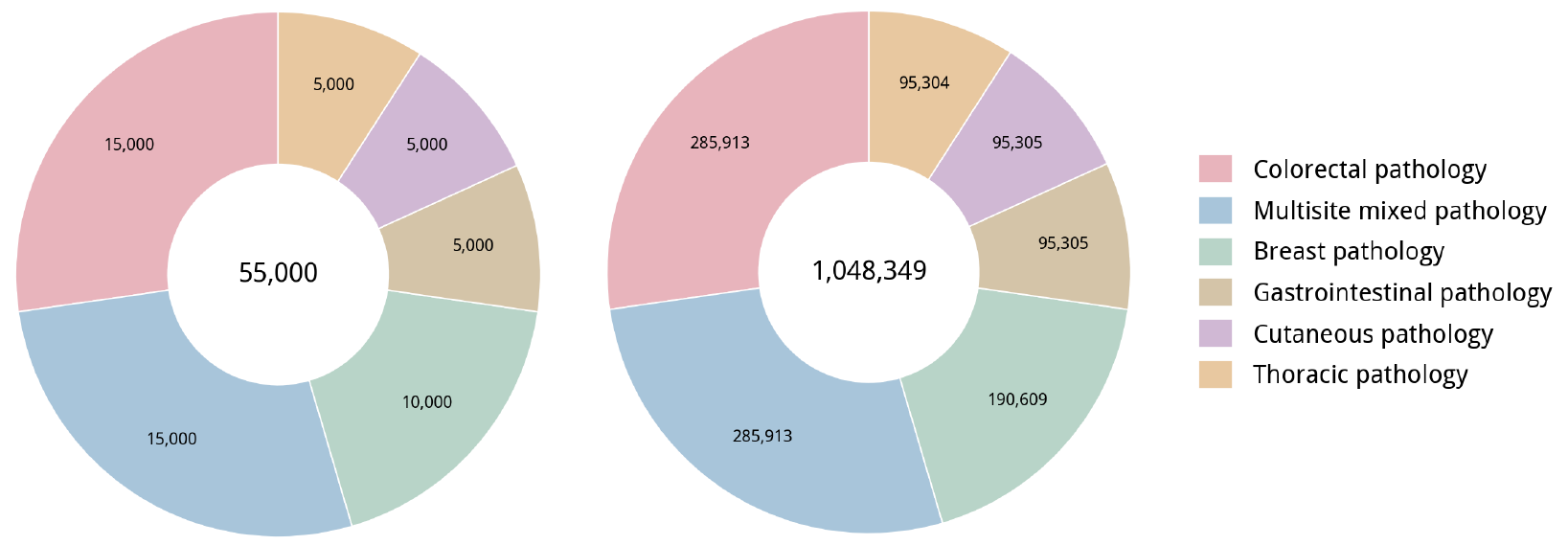}
\caption{Organ-source distribution of unlabeled pathology images used for masked-diffusion pretraining. The left donut chart shows the 55,000-image method-development corpus, and the right donut chart shows the 1,048,349-image large-scale experimental corpus.}
\label{fig:pretrain_organ_distribution}
\end{figure}

We use unlabeled pathology patches sourced from the public HistAI collection \cite{nechaev2025histai} for masked-diffusion pretraining and evaluate the learned frozen representations on downstream cell-level dense prediction datasets. Figure~\ref{fig:pretrain_organ_distribution} summarizes the organ-source distribution of the unlabeled pretraining corpus. We strictly confirm that the pretraining dataset contains no overlapping samples with downstream evaluation data to eliminate data leakage concerns.
\FloatBarrier

\subsection{ConvNeXt Masked-Diffusion Architecture Details}
\label{app:cmd_architecture}

Table~\ref{tab:cmd_architecture} summarizes the architectural differences between CMD-B and CMD-L, together with their shared ConvNeXt masked-diffusion design. The two variants use the same U-shaped topology and differ mainly in model width. CMD-L increases the number of channels at each stage, providing higher representation capacity while preserving the same resolution hierarchy, block allocation, conditioning mechanism, and downstream feature extraction protocol.

\begin{table}[t]
\centering
\caption{Architecture details of CMD-B and CMD-L. Variant-specific settings are listed at the top, while shared architectural components are grouped below.}
\label{tab:cmd_architecture}
\small
\setlength{\tabcolsep}{5pt}
\renewcommand{\arraystretch}{1.15}
\begin{tabular}{p{0.28\linewidth}p{0.29\linewidth}p{0.29\linewidth}}
\toprule
\textbf{Component} & \textbf{CMD-B} & \textbf{CMD-L} \\
\midrule
Base channel width
& 256
& 512 \\

Channel hierarchy
& $256,256,256,512,512,1024$
& $512,512,512,1024,1024,2048$ \\

Trainable parameters
& $\sim$130.75M
& $\sim$516.71M \\

\midrule
\multicolumn{3}{l}{\textbf{Shared ConvNeXt Masked-Diffusion Design}} \\
\midrule
Input/output
& \multicolumn{2}{p{0.58\linewidth}}{RGB pathology image input and RGB reconstruction target at $256 \times 256$ resolution.} \\

Backbone topology
& \multicolumn{2}{p{0.58\linewidth}}{ConvNeXt-style U-Net with five downsampling stages, a bottleneck, and five mirrored decoder stages.} \\

Resolution hierarchy
& \multicolumn{2}{p{0.58\linewidth}}{$256^2 \rightarrow 128^2 \rightarrow 64^2 \rightarrow 32^2 \rightarrow 16^2 \rightarrow 8^2$, followed by symmetric upsampling back to $256^2$.} \\

Encoder depth
& \multicolumn{2}{p{0.58\linewidth}}{The five encoder stages contain $1,2,3,2,2$ ConvNeXt blocks, respectively.} \\

Bottleneck depth
& \multicolumn{2}{p{0.58\linewidth}}{The lowest-resolution $8 \times 8$ bottleneck contains 6 ConvNeXt blocks for global tissue-context aggregation.} \\

Decoder depth
& \multicolumn{2}{p{0.58\linewidth}}{The five decoder stages contain $2,2,3,2,1$ ConvNeXt blocks, respectively.} \\

Total ConvNeXt blocks
& \multicolumn{2}{p{0.58\linewidth}}{26 blocks in total.} \\

ConvNeXt block
& \multicolumn{2}{p{0.58\linewidth}}{$7 \times 7$ depthwise convolution, LayerNorm, pointwise MLP, GELU, GRN, residual connection, and adaptive modulation.} \\

MLP expansion ratio
& \multicolumn{2}{p{0.58\linewidth}}{3.} \\

Down/up sampling
& \multicolumn{2}{p{0.58\linewidth}}{Pixel-unshuffle downsampling and pixel-shuffle upsampling.} \\

Skip connection
& \multicolumn{2}{p{0.58\linewidth}}{Encoder features are concatenated with decoder features at the same resolution and compressed by a $1 \times 1$ convolution.} \\

Conditioning
& \multicolumn{2}{p{0.58\linewidth}}{Diffusion timestep embedding and frozen pathology feature are projected to each stage dimension and fused by addition.} \\

Condition injection
& \multicolumn{2}{p{0.58\linewidth}}{Adaptive LayerNorm-Zero modulation is applied in every ConvNeXt block.} \\

Feature extraction
& \multicolumn{2}{p{0.58\linewidth}}{During downstream dense prediction, the reconstruction output is discarded and multi-scale decoder features are reused as frozen dense representations.} \\
\bottomrule
\end{tabular}
\end{table}

\subsection{Downstream Segmentation Head}
\label{app:downstream_head}

\begin{figure}[htbp!]
\centering
\includegraphics[width=\linewidth]{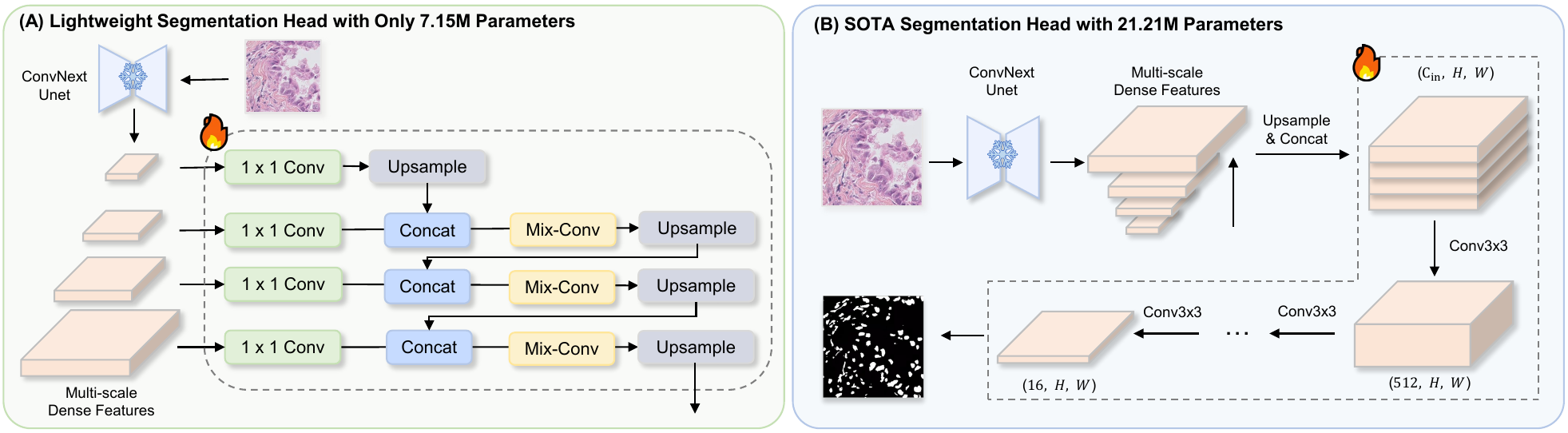}
\caption{Two downstream segmentation heads built on frozen CMD dense features. (A) The lightweight segmentation head has 7.15M trainable parameters and fuses multi-scale decoder features in a top-down manner. (B) The SOTA-level segmentation head has 21.21M trainable parameters and uses a convolutional decoder after multi-scale feature concatenation.}
\label{fig:downstream_heads}
\end{figure}

After pretraining, the CMD diffusion backbone is frozen and used as a dense feature extractor. We evaluate two downstream segmentation heads, as illustrated in Fig.~\ref{fig:downstream_heads}. The first is a lightweight segmentation head with 7.15M trainable parameters, corresponding to the setting reported in Table~\ref{tab:fewshot_performance}. The second is a stronger SOTA-level segmentation head with 21.21M trainable parameters, also reported in Table~\ref{tab:fewshot_performance}.

For the lightweight head in Fig.~\ref{fig:downstream_heads}(A), we follow the visual encoder design of Li et al.~\cite{li2023open}. Multi-scale decoder features are extracted from selected frozen diffusion decoder stages. Each feature is first projected to a unified dimension $d=256$ by a $1 \times 1$ convolution. Starting from the coarsest scale, features are progressively upsampled, concatenated with the next finer feature map, and blended by a Mix-Conv module. Each Mix-Conv contains two $3 \times 3$ convolutions with a residual connection and conditional batch normalization. The final fused feature is passed to a lightweight $1 \times 1$ convolutional segmentation head.

For the SOTA-level head in Fig.~\ref{fig:downstream_heads}(B), dense features from selected decoder blocks are resized to the target segmentation resolution and concatenated along the channel dimension, producing a feature map of shape $(C_{\text{in}}, H, W)$. In the FCDM-L setting, $C_{\text{in}}=4096$ using decoder blocks $[1,3,5,6,8,9]$ at diffusion step $t=50$. The head first applies a $3 \times 3$ convolution with batch normalization and ReLU to reduce the channel dimension to 512, followed by four convolutional decoder blocks with output channels 256, 128, 64, and 16. Each block contains two $3 \times 3$ Conv-BN-ReLU layers and preserves spatial resolution. A final $3 \times 3$ convolution maps the 16-channel feature map to segmentation logits.

\subsection{Hyperparameters and Implementation Details of Method-Section}
\label{app:method_config}

We summarize the configuration used in our method-section. 
Table~\ref{tab:backbone_config} details the architectural specifications of the compared backbones. 
To keep the comparison controlled, ConvNeXt-U-Net, Attention U-Net, and DiT are trained with the same masked-diffusion objective and the same frozen pathology foundation model conditioning, with unified pretraining and downstream training settings summarized in Table~\ref{tab:maskdiff_training_config}. 
The ConvNeXt-U-Net model is our default backbone, while Attention U-Net-B and DiT-B serve as architecture ablations.

\begin{table}[t]
\centering
\caption{Backbone configurations for method-section masked-diffusion models.}
\label{tab:backbone_config}
\small
\setlength{\tabcolsep}{4pt}
\renewcommand{\arraystretch}{1.15}
\begin{tabular}{p{0.24\linewidth}p{0.23\linewidth}p{0.23\linewidth}p{0.20\linewidth}}
\toprule
\textbf{Configuration} & \textbf{ConvNeXt-U-Net} & \textbf{Attention U-Net} & \textbf{DiT} \\
\midrule

Backbone type
& ConvNeXt-style U-Net
& Hybrid-style U-Net
& Transformer \\

Default input space
& Pixel space
& Pixel / VAE latent
& VAE latent \\

Input size
& $256 \times 256$
& $256 \times 256$
& $256 \times 256$ \\

Parameters
& $\sim$130.75M
& $\sim$132.65M
& $\sim$131M \\

\bottomrule
\end{tabular}
\end{table}

\begin{table}[t]
\centering
\caption{Shared masked-diffusion and downstream training settings of Method-Section.}
\label{tab:maskdiff_training_config}
\small
\setlength{\tabcolsep}{5pt}
\renewcommand{\arraystretch}{1.12}
\begin{tabular}{p{0.34\linewidth}p{0.58\linewidth}}
\toprule
\textbf{Configuration} & \textbf{Setting} \\
\midrule
Pretraining data & 55,000 unlabeled pathology images \\
Diffusion timesteps & $T=1000$ \\
Pixel-space mask & $8 \times 8$ patches, producing a $32 \times 32$ mask grid \\
VAE-space mask & $4 \times 4$ latent patches, producing an $8 \times 8$ mask grid \\
Optimizer & AdamW, weight decay 0 \\
Pretraining schedule & 50,000 steps, learning rate $3\times10^{-5}$ \\
EMA & 0.9999 \\
Pathology conditioning & Frozen UNI feature, 1024 dimensions \\
VAE variant & SD-VAE with 4 latent channels and scaling factor 0.18215 \\
Downstream protocol & Linear Probe \\
Downstream  timestep & $t=50$ \\
Downstream training & 150 epochs, AdamW, learning rate $10^{-3}$, cosine schedule \\
Downstream loss & Cross-entropy plus Dice loss \\
\bottomrule
\end{tabular}
\end{table}

\section{Comparison with Test-Time Training}
\label{app:ttt}
\renewcommand{\thefigure}{C\arabic{figure}}
\renewcommand{\thetable}{C\arabic{table}}
\setcounter{figure}{0}
\setcounter{table}{0}

We further compare CMD-L with Test-Time Training (TTT) on CPM-17 and TNBC. TTT relies on test-time optimization, while CMD-L uses a frozen pretrained diffusion backbone and performs standard feed-forward inference with downstream segmentation heads. As shown in Table~\ref{tab:ttt_cmd_comparison}, CMD-L achieves performance close to TTT without test-time adaptation, \textbf{highlighting the effectiveness of the learned frozen dense representation}.

\begin{table}[htbp]
\centering
\caption{Comparison between Test-Time Training (TTT) and CMD-L on CPM-17 and TNBC. CMD-L is evaluated with two downstream segmentation heads from Fig.~\ref{fig:downstream_heads}. Results are reported as mean Dice/Precision with 95\% confidence intervals.}
\label{tab:ttt_cmd_comparison}
\footnotesize
\setlength{\tabcolsep}{3pt}
\begin{tabular}{l cc cc}
\toprule
\multirow{2}{*}{\textbf{Method}} & \multicolumn{2}{c}{\textbf{CPM-17}} & \multicolumn{2}{c}{\textbf{TNBC}} \\
\cmidrule(lr){2-3} \cmidrule(lr){4-5}
& Dice & Precision & Dice & Precision \\
\midrule

\shortstack[l]{\textbf{CMD-L(ours)}\\\footnotesize + SegHead: Fig.~\ref{fig:downstream_heads}(A)}
& \vc{0.865}{0.827}{0.902}
& \vc{0.901}{0.876}{0.925}
& \vc{0.854}{0.824}{0.891}
& \vc{0.851}{0.822}{0.876} \\

\shortstack[l]{\textbf{CMD-L(ours)}\\\footnotesize + SegHead: Fig.~\ref{fig:downstream_heads}(B)}
& \vc{0.880}{0.847}{0.908}
& \vc{0.883}{0.860}{0.920}
& \vc{0.854}{0.835}{0.874}
& \vc{0.830}{0.802}{0.858} \\

\shortstack[l]{\textbf{CMD-L(ours)} + TTT\\\footnotesize + SegHead: Fig.~\ref{fig:downstream_heads}(B)} 
& \vc{0.885}{0.862}{0.908} 
& \vc{0.892}{0.866}{0.924} 
& \vc{0.861}{0.842}{0.880} 
& \vc{0.860}{0.823}{0.894} \\
\bottomrule
\end{tabular}
\end{table}



\end{document}